%% file: main.tex
\documentclass[10pt,twocolumn,letterpaper]{article}

\usepackage{cvpr}      
\usepackage{bbding}
\usepackage{array} 
\usepackage{multicol}
\usepackage{multirow}
\input{preamble}

\definecolor{cvprblue}{rgb}{0.21,0.49,0.74}
\usepackage[pagebackref,breaklinks,colorlinks,citecolor=cvprblue]{hyperref}
\usepackage{tabularx}
\usepackage{makecell}
\usepackage[pagebackref,breaklinks,colorlinks,citecolor=cvprblue]{hyperref}
\usepackage{tabularx}
\usepackage{makecell}

\usepackage{algorithm}
\usepackage{algorithmic}
\usepackage{verbatim}
\usepackage{algorithm}
\usepackage{algorithmic}
\usepackage{verbatim}

\def\paperID{5675} 
\def\confName{CVPR}
\def\confYear{2024}

\title{Boosting Order-Preserving and Transferability for Neural Architecture Search: a Joint Architecture Refined Search and Fine-tuning Approach}

\author{\textbf{Beichen Zhang},~\textbf{Xiaoxing Wang},~\textbf{Xiaohan Qin},~\textbf{Junchi Yan}\thanks{Correspondence author. The work was partly supported by NSFC (92370201, 62222607).}\\
{Department of Computer Science and Engineering \& MoE Key Lab of AI, Shanghai Jiao Tong University}\\
{\tt \{zhangbeichen,figure1\_wxx,galaxy-1,yanjunchi\}@sjtu.edu.cn}\\
\small \url{https://github.com/beichenzbc/Supernet-shifting}}

\begin{document}
\maketitle

\begin{abstract}

Supernet is a core component in many recent Neural Architecture Search (NAS) methods. It not only helps embody the search space but also provides a (relative) estimation of the final performance of candidate architectures. Thus, it is critical that the top architectures ranked by a supernet should be consistent with those ranked by true performance, which is known as the order-preserving ability.
In this work, we analyze the order-preserving ability on the whole search space (global) and a sub-space of top architectures (local), and empirically show that the local order-preserving for current two-stage NAS methods still need to be improved. To rectify this, we propose a novel concept of \textbf{Supernet Shifting}, a refined search strategy combining architecture searching with supernet fine-tuning. Specifically, apart from evaluating, the training loss is also accumulated in searching and the supernet is updated every iteration. Since superior architectures are sampled more frequently in evolutionary searching, the supernet is encouraged to focus on top architectures, thus improving local order-preserving.
Besides, a pre-trained supernet is often un-reusable for one-shot methods. We show that Supernet Shifting can fulfill transferring supernet to a new dataset. Specifically, the last classifier layer will be unset and trained through evolutionary searching. Comprehensive experiments show that our method has better order-preserving ability and can find a dominating architecture. Moreover, the pre-trained supernet can be easily transferred into a new dataset with no loss of performance. 

\end{abstract}
\section{Introduction}
\label{sec:intro}
Apart from traditional manually designed neural networks e.g. VGG~\cite{vgg} and RESNET~\cite{resnet}, Neural Architecture Search (NAS), as an important part of Automated Machine Learning (AutoML), aims to automatically search an optimal architecture in a certain search space.


Early NAS approaches~\cite{ref,ref2,ref3,ref4} adopt a time-consuming pipeline by sampling architectures and training their weights separately.
To speed up the search procedure, many recent approaches introduce performance estimators for each architecture. Though there exists a performance estimation gap, the search stage focuses more on the relative ranking of architectures rather than true performance.

Some works~\cite{grasp,snip,naswot} propose zero-cost proxies, which only require one forward step or one backward step. Nevertheless, it cannot perform consistently well on diverse tasks~\cite{eznas}. Other works~\cite{ENAS,oneshot,FAIR} build up a supernet that contains all candidate operations and connections in the search space. Once a supernet is trained, all architectures can be quickly evaluated by loading the supernet weight. 
\begin{figure}[tb!]
  \centering
   \includegraphics[width=0.98\linewidth]{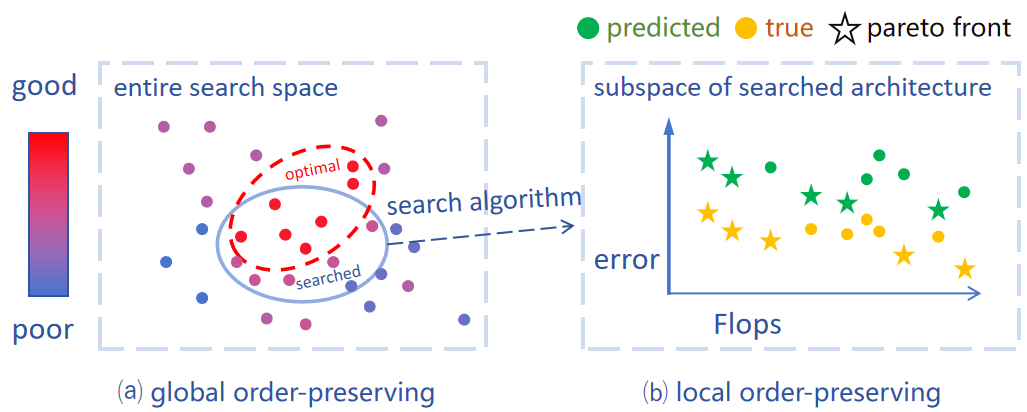}
   \vspace{-10pt}
   \caption{An illustration of global and local order-preserving ability. For global one, we care about coarse-grained comparison to wipe out poor architectures in entire search space. For local one, we care about fine-grained comparison to rank the architectures in a subspace of top architectures.}
   \label{fig:global-local}
\end{figure}
Since such a supernet-based performance estimator is widely used, the order-preserving ability, i.e. whether the estimated architecture ranking is consistent with true performance ranking, is an important index. 
In this work, we delve into two aspects of order-preserving ability.
One is \textbf{global order-preserving ability} to distinguish good architectures from poor ones in the whole search space. The other is \textbf{local order-preserving ability} to rank top architectures that have good performance.
We argue that both the above abilities are important. Poor global order-preserving abilities will lead to unsatisfactory results since it cannot screen out poor architectures. Poor local order-preserving ability will reduce search efficiency.





Some NAS approaches, like DARTS~\cite{DARTS} and GDAS~\cite{GDAS}, introduce an architecture parameter optimized during supernet training. It lets supernet focus on part of the architectures, improving local resolution. However, it can't ensure global order-preserving ability. Some previous works~\cite{SPOS,drop,partial} have pointed out that an early bias is easily introduced. The search direction may be led wrongly in the early stage.

Other works like SPOS~\cite{SPOS} and FairNAS~\cite{FAIR} ensure global order-preserving ability. They treat NAS as a two-stage process. First, a single-path supernet is trained by uniform sampling and all architectures' weights are optimized equally. Then, a searching algorithm like evolutionary algorithm (\textbf{EA}) is applied. This method ensures fairness in the training stage so global order-preserving ability is better.
However, the local order-preserving ability still needs to be improved. We select the top-10 architectures during searching and retrain them separately. The \emph{Kendall's tau} of the accuracy in supernet and after retrain is only 0.17, which is far from satisfactory. While although the choice blocks are similar, there do exist a non-negligible accuracy gap of 0.8\% after retrain. The problem is that weights of different architectures are not fully decoupled due to weight-sharing strategy. Some works~\cite{SUMNAS,m2,m3,m4,m5} point out that the phenomenon of multi-model forgetting exists in weight sharing. Uniform sampling strategy avoids early bias and ensures fairness, but may not be precise in local comparison. In fact, the supernet is encouraged to focus on some local superior architectures. Inferior architectures may produce noise and should be neglected.

However, ensuring both isn't straightforward. Uniform sampling isn't precise enough. Biased sampling is needed to focus on some superior architectures, but it can lead to improper bias and hurt global order-preserving ability.

To this end, we propose a refined search strategy combining architecture searching with supernet fine-tuning to both achieve high global and local order-preserving ability. 
To ensure global order-preserving, we first train a supernet by uniform sampling to avoid early bias and ensure fairness. To improve local order-preserving, we add a \textbf{Supernet Shifting} stage during searching. Specifically, we calculate and accumulate the training loss together with evaluation when an architecture is sampled. After each iteration of evolutionary searching, the supernet is updated. Since superior architectures are sampled more frequently in evolutionary searching, our shifted supernet is encouraged to focus on top architectures, thus having better local discernment and better local order-preserving ability. While for inferior architectures, the supernet gradually forgets them. Unlike superior architectures, we don't care about the fine-grained estimation since they should all be eliminated in searching. 

Moreover, our method has better transferability compared to other one-shot NAS methods that have to train a new supernet for a new dataset. In contrast, our method can adopt the original supernet and fine-tune the weight during the search procedure. Plenty of previous work on transfer learning and pre-trained models~\cite{bert,MAE} proves that weights can be inherited and only need some small changes for different tasks. This provides a theoretical basis to transfer supernet to different datasets.

Overall, our contributions can be summarized as follows:





\textbf{1) Comprehensive analysis about order-preserving ability for NAS methods.}
Many NAS methods adopt proxies to estimate performance to speed up the search process, making order-preserving ability a general metric. In this work, we further define global and local order-preserving ability and verify the dilemma of current NAS method. 

\textbf{2) A stable strategy improving both global and local order-preserving ability.}
This work introduces supernet shifting, a simple but effective method to improve global and local order-preserving ability. Supernet is self-adaptively revised during searching. Compared with previous method, our method is by design more in line with the essence of NAS in the sense of paying adaptive attention to different architectures, and meanwhile introduces less bias.

\textbf{3) A flexible and efficient strategy realizing transferring during searching.}
 Our method neatly realizes the transfer of supernet by reusing the feature encoder part of pre-trained supernet. This allows flexible design of decoders to better fit different tasks. Transferring with searching further ensures efficiency. To the best of our knowledge, this is the first time an entire supernet can be transferred.
 
\textbf{4) Strong Performance.}
Experiments in Sec.~\ref{sec:4} show the general effectiveness of our method. It improves both global and local order-preserving ability and can obtain dominating architectures. Flops can be reduced by $5M$ and the accuracy increases by $0.3\%$ on ImageNet-1K. For transferability, it can accelerate searching process for ten times compared with SPOS without performance loss.  


\section{Related Works}
\label{sec:formatting}

\textbf{One-shot NAS.} 
Early NAS approaches~\cite{ref,ref2,ref3,ref4} train different architectures separately. The cost of time is often unaffordable.
\emph{ENAS} ~\cite{ENAS} introduces weight-sharing strategy so that different architectures can be jointly optimized. This greatly speeds up the NAS process.
Based on weight-sharing strategy, \emph{One-Shot NAS}~\cite{oneshot} further adopts path dropout technique to train a supernet. Once a supernet is trained, the architectures can simply inherit weight from it. Most follow-up works adopt this strategy. They can be roughly divided into two classes based on their search space.
Some works ~\cite{DARTS,GDAS,pdarts,proxy,dense} 
adopt contiguous search space. They introduce an architecture parameter and use gradient decent to optimize the supernet weight and the architecture parameter.
Another line of works ~\cite{SPOS,FAIR,single} adopt discrete search space. They treat NAS as a two-stage problem. In supernet training stage, only one path is sampled and optimized from search space. Then searching algorithms are applied to find the optimal architecture. This type of method is usually more stable and is easier to optimize.
A widely used NAS method \emph{SPOS}~\cite{SPOS} falls into the second category. \emph{SPOS} trains a supernet by uniform sampling and applies evolutionary searching algorithm to search for optimal architecture. This is our main baseline.

\textbf{Evolutionary-based NAS.} 
Evolutionary algorithm is widely used in discrete optimization problem. It simulates the process of biological evolution process. New candidates are created by crossover and mutation.
\emph{Large-Scale Evolution of Image Classifiers} ~\cite{ea1} first introduces evolutionary algorithm into NAS problem. New candidate architectures are produced and trained separately. 
After the proposal of weight sharing strategy and supernet, evolutionary algorithm is widely used in the works ~\cite{SPOS,FAIR} using discrete search space as a sample strategy during searching stage. Specifically, \emph{SPOS}~\cite{SPOS} applies evolutionary searching algorithm first on a single-path supernet while \emph{FairNAS}~\cite{FAIR} uses NSGA-\uppercase\expandafter{\romannumeral2}. 
It's widely accepted that evolutionary searching algorithm is more efficient than random searching since it takes use of the current evaluation results.

\textbf{Transferability in NAS.} 
Transfer learning aims to find an effective way to transfer the knowledge learned from a source domain to a target domain. Usually, the weights of the pre-trained model are carefully fine-tuned and the structure is often adjusted to adapt it to the target domain.
Transferability is an important feature for NAS method, since it can significantly reduce the time cost and hardware restriction for NAS application. However, few NAS methods find an effective way to realize tranferability. Some NAS methods~\cite{DARTS,SNAS} first search on a small-scale dataset like Cifar10~\cite{cifar}, then extend the searched architecture to large-scale datasets like ImageNet-1K~\cite{image}. This transfer mode could be unreliable because of the diverse data distribution. Some works~\cite{trans, grouped} introduce meta learning to achieve transferability. However, this is often complicated. The model and hyperparameters have to be carefully designed and they usually lack explainability.
To the best of our knowledge, no previous work has transferred an entire pre-trained supernet to a new dataset. 


\begin{figure}[tb!]
  \centering
   \includegraphics[width=0.98\linewidth]{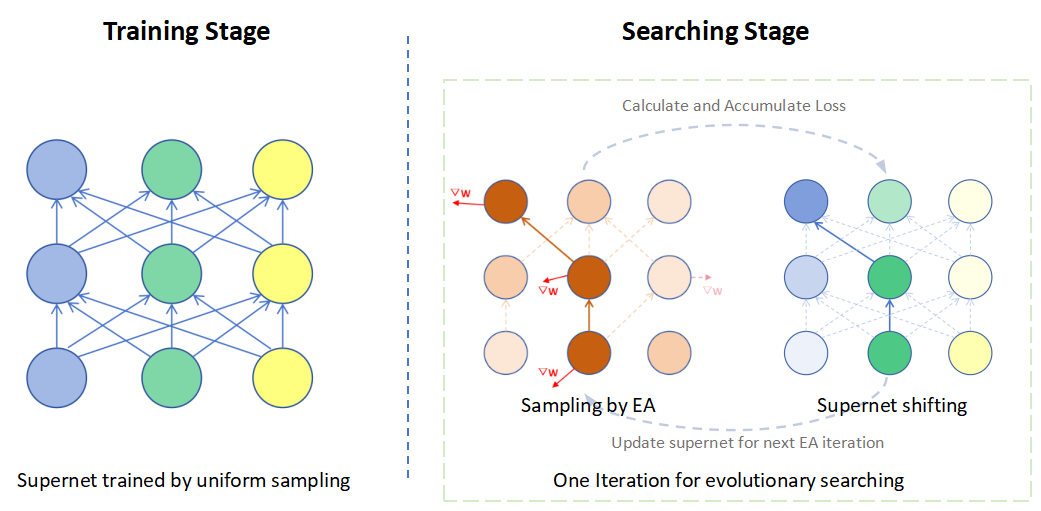}
   \vspace{-10pt}
   \caption{Pipeline of our method with two stages. In the training stage, a single-path supernet is trained by uniform sampling. Each architecture is equally treated. In the searching stage, evolutionary searching is applied. When an architecture is sampled, the training loss is calculated and accumulated apart from evaluating. At the end of each iteration, the supernet is updated. Since superior architectures are sampled more frequently in evolutionary searching, the supernet is expected to shift to focus on top architectures.}
   \label{fig:overall}
\end{figure}

\begin{table}[tb!]
  \centering
  \caption{Overview of different one-shot NAS methods. By adding supernet shifting, our method introduces extra attention on partial architectures based on their validation performance. Therefore, the local resolution can be improved and less bias is introduced. Besides, ours has the best transferability.}
  \vspace{-5pt}
  \large
  \resizebox{0.47\textwidth}{!}
{
  \begin{tabularx}{\textwidth}{|>
  {\centering\arraybackslash}m{2.15cm}<{\centering}|>{\centering\arraybackslash}m{2.85cm}<{\centering}|>{\centering\arraybackslash}m{1.45cm}<{\centering}|>{\centering\arraybackslash}m{3.95cm}<{\centering}|>{\centering\arraybackslash}m{0.98cm}<{\centering}|>
  {\centering\arraybackslash}m{0.96cm}<{\centering}|>
  {\centering\arraybackslash}m{2.06cm}<{\centering}|}
    \hline
     \textbf{Method} & \textbf{Supernet training} & \textbf{Search} & \makecell[c]{\textbf{Attention to}\\ \textbf{different architectures}} & 
     \textbf{global resol.} &
    \textbf{local resol.}&
    \textbf{transferable supernet}\\
    \hline
    DARTS~\cite{DARTS} & gradient descent, path dropout & / & \makecell[c]{biased by architecture\\ parameter at start}  &\XSolidBrush &\Checkmark & \XSolidBrush \\
    \hline
    OneShot~\cite{oneshot} & path dropout & random search & equal attention  &\Checkmark &\XSolidBrush &\XSolidBrush \\ 
    \hline
    SPOS~\cite{SPOS} & uniform sampling & EA & equal attention &\Checkmark &\XSolidBrush  &\XSolidBrush \\
    \hline
    FairNAS~\cite{FAIR} & strict fairness sampling & EA &
    strictly equal attention&\Checkmark & \XSolidBrush & \XSolidBrush \\
    \hline
    Ours & uniform sampling & EA with shifting &\makecell[c]{biased by validation\\ after uniform training} & \Checkmark &\Checkmark &\Checkmark\\
    \hline 
  \end{tabularx}}
  \label{tab:overview}
\end{table}

\section{Method}
\subsection{NAS Retrospection from Order-preserving}
To have a deeper understanding of different NAS methods and their advantages as well as shortcomings, we use some mathematical expressions to show their training pipelines and training goals. Thus, we need to make some definitions for the following mathematical expressions.

We denote $\mathcal{N}$ as the supernet, $\theta$ as the learned architecture variable, $\alpha$ represents different architectures,  $\tau(\alpha)$ represents a sampling distribution of architecture $\alpha$. $\mathcal{A}$ represents the entire search space and $u(\mathcal{A})$ represents a uniform sampling over the search space. $\mathcal{W_{\mathcal{A}}}$ represents the weight of the entire supernet and $\mathcal{W}^{\alpha}_{\mathcal{A}}$ represents the weight of architecture $\alpha$ inherited from supernet $\mathcal{W_{\mathcal{A}}}$. 

As discussed above, we divide the previous one-shot NAS into two main categories. Methods in the first category~\cite{DARTS,GDAS,pdarts,proxy,dense} use continuous search space and introduce an architecture variable $\theta$ which is jointly optimized with supernet weights. Their optimization steps are described as follows:
\begin{equation}
  (\theta, \mathcal{W}_{\theta}) = \mathop{\arg\min}\limits_{\theta, \mathcal{W}}\mathcal{L}_{train}(\mathcal{N}(\mathcal{A}(\theta),\mathcal{W}))
  \label{eq:4}
\end{equation}

The existence of $\theta$ leads the supernet to focus on a local set of architectures, and this improves the local resolution near $\theta$. This kind of method theoretically works well if $\theta$ is optimal. However, as some works~\cite{SPOS,drop,partial} points out, $\theta$ is jointly optimized and this may introduce early bias. So 
the supernet may be misled and trapped into local optimal architecture, which hurts global order-preserving ability.

Approaches in the other category~\cite{SPOS,FAIR,single} use discrete search space and get rid of the learnable architecture variable $\theta$. They first train a supernet by uniform sampling and apply searching algorithms to find the optimal architecture based on their performance on the supernet. Their optimization steps are sequential, which can be illustrated as:
\begin{equation}
  \mathcal{W_{\mathcal{A}}} = \mathop{\arg\min}\limits_{\mathcal{W}}\mathbb{E}_{\alpha \sim u(\mathcal{A})}[\mathcal{L}_{train}(\mathcal{N}(\alpha,\mathcal{W}))]
  \label{eq:5}
\end{equation}

\begin{equation}
  \alpha^{*} = \mathop{\arg\max}\limits_{\alpha}ACC_{val}(\mathcal{N}(\alpha,\mathcal{W}^{\alpha}_{\mathcal{A}}))
  \label{eq:51}
\end{equation}

By removing $\theta$ and applying uniform sampling, it better achieves the fairness of comparison and the global order-preserving ability. However, it isn't precise enough for local comparison because it pays equal attention to every single architecture. Our work aims to achieve both global and local order-preserving ability. Sec.~\ref{sec:3.2} aims to ensure the former and Sec.~\ref{sec:3.3} aims to improve the latter.

\subsection{Single-Path Supernet Training}
\label{sec:3.2}
First, we construct a single-path supernet. The construction and training of supernet follow SPOS~\cite{SPOS}. Specifically, the supernet has a series of choice blocks, and each choice block further could be specified by certain configurations. Only one choice is invoked at the same time. The supernet is trained by uniform sampling. Every single architecture is treated equally. The training process can be defined as:
\begin{equation}
  \mathcal{W_{\mathcal{A}}} = \mathop{\arg\min}\limits_{\mathcal{W}}\mathbb{E}_{\alpha \sim u(\mathcal{A})}[\mathcal{L}_{train}(\mathcal{N}(\alpha,\mathcal{W}))]
  \label{eq:6}
\end{equation}

As will be shown in the experiments in Sec.~\ref{sec:4.3}, global order-preserving ability for the supernet trained by uniform sampling is quite ideal, so that most poor architectures can be effectively eliminated. However, the local order-preserving ability is still far from satisfactory.

\subsection{Supernet Shifting}
\label{sec:3.3}
After training a supernet by uniform sampling, the variance of different architectures' accuracy is relatively small. This is enough for global comparison. The supernet can distinguish between good architecture and poor architecture. However, it's not capable enough to distinguish the best architecture from several similar architectures, and this seriously affects its performance. 

To solve the problem, the supernet weights should be shifted based on the performance of different architectures evaluated on the current supernet. Specifically, the architectures which perform better should be more emphasized and sampled more frequently. It can be depicted as:
\begin{equation}
  \mathcal{W_{\mathcal{A}^{*}}} = \mathop{\arg\min}\limits_{\mathcal{W}}\mathbb{E}_{\alpha \sim\tau(\mathcal{\alpha})}[\mathcal{L}_{train}(\mathcal{N}(\alpha,\mathcal{W}^{\alpha}_{\mathcal{A}})]
  \label{eq:7}
\end{equation}

The prior distribution $\tau(\alpha)$ is important. If $\tau(\alpha)$ is a uniform, the equation is the same as Eq.~\ref{eq:6}. As discussed above, for architecture $\alpha$, we want the sampling probability $\tau(\alpha)$ and the performance of $\alpha$ to be positively correlated:
\begin{equation}
r[\tau({\alpha}),ACC_{val}(\mathcal{N}(\alpha,\mathcal{W}^{\alpha}_{\mathcal{A}}))]\ > \ 0
  \label{eq:71}
\end{equation}

Thus, after the shifting stage, the supernet is led to focus on better architectures. We then search for the optimal architecture based on the shifted supernet.
\begin{equation}
  \alpha^{*} = \mathop{\arg\max}\limits_{\mathcal{\alpha}}ACC_{val}(\mathcal{N}(\alpha,\mathcal{W}^{\alpha}_{\mathcal{A}^{*}}))
  \label{eq:8}
\end{equation}

Compared with those biased architecture sampling strategies using architecture parameter $\theta$~\cite{GDAS,DARTS,pdarts}, this sampling strategy could be more precise and more reliable as fairness can be ensured in the supernet training stage and the sampling distribution is determined by their evaluation performance directly.

Till now, there is problem remaining unsolved. A proper sampling strategy satisfying Eq.~\ref{eq:71} isn't straightforward.

We take use of the property of evolutionary searching algorithm to solve the problem. The main difference between evolutionary searching algorithm and random searching is that, for different architectures, the probability of being sampled by evolutionary searching algorithm is different. The evolutionary searcher tends to sample the architectures which are similar to the current top architectures, so better architectures are sampled more frequently. Therefore, we can claim that the sampling probability of evolutionary searching algorithm satisfies Eq.~\ref{eq:71}
\begin{figure}[t!]
  \centering
   \includegraphics[width=0.98\linewidth]{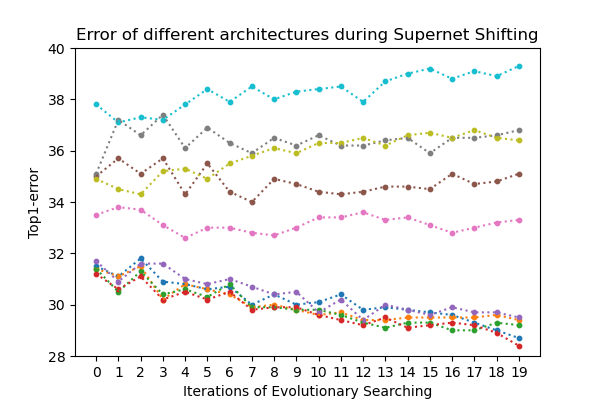}
   \vspace{-5pt}
   \caption{Trajectory of Supernet Shifting process. We sample 5 searched superior architectures (bottom) and 5 random architectures (top). We monitor their error rate over iterations of evolutionary searching. Iteration 0 denotes the original supernet trained by uniform sampling. The shifting supernet gradually focuses on superior architectures and dismisses inferior ones.}
   \label{fig:shifting}
\end{figure}

Thus, we implement Eq.~\ref{eq:7} and Eq.~\ref{eq:8} homogeneously and realize supernet shifting together with evolutionary searching. Specifically, when an architecture is sampled in evolutionary searching stage, apart from evaluating the validation performance, the loss is calculated according to tens of training iterations and is accumulated. After one iteration of evolutionary searching, the supernet is updated. Thus, it will be shifted to focus on a local set of good architectures due to the biased sampling distribution. This improves the precision for local comparison.

Fig.~\ref{fig:shifting} shows the supernet shifting process during the evolutionary searching stage. The accuracy increases for superior architectures while remains or even decreases for inferior ones. This verifies our assumption.

Since supernet keeps shifting in searching, we have to eliminate the bias produced by the changing supernet. Therefore, repeatedly sampling should be allowed. Specifically, for a single particular architecture, we allow a new sample in each iteration of evolutionary algorithm and we will update its latest accuracy. After each iteration, we keep the top-50 different architectures for later mutation and crossover process.

As will be empirically shown in the experiment in Sec.~\ref{sec:4.2} and Sec.~\ref{sec:4.3}, adding the additional shifting stage significantly improves the local order-preserving ability on the superior architectures searched by evolutionary searching algorithm and the global order-preserving ability doesn't decay. 
One can obtain dominating architectures on different datasets by applying supernet shifting. This verifies the general effectiveness of our method.

Since we implement supernet shifting and architecture searching homogeneously, and the shifting process need few iterations for training comparing with evaluation, the additional time cost is quite little. Take ImageNet-1K as an example, the total time for evolutionary searching stage only rises from 17 GPU hours to 19 GPU hours.

Alg.~\ref{alg:supernet_shifing} shows the entire searching stage of our method. It can be applied in any two-stage NAS method, after a supernet is trained to improve the local resolution of the supernet.

\begin{algorithm}[tb!]
\caption{Supernet shifting during EA searching} \label{alg:supernet_shifing}
\begin{algorithmic}[1]
\REQUIRE Pre-trained Supernet $\mathcal{N}$ with weights $\mathcal{W}$, Size of a population $T$.
\ENSURE Pareto optimal architectures.
\STATE Initialize a population of architectures $Top\_T$ and get the validation accuracy
\REPEAT
    \STATE Generate a new population of architectures by EA $\{\mathcal{A}_t\}_{t=1}^T$ = $Generate\_new\_candidates(Top\_T, T)$
    \STATE Initialize total gradient $\nabla\mathcal{W}=0$
\FOR {$t = 0$ to $T$}
    
    \STATE Get validation accuracy for architecture $\mathcal{A}_t$ $ACC_{\mathcal{A}_t} = validate(\mathcal{N}, \mathcal{W}, \mathcal{A}_t, val\_data)$
    
    \STATE compute and accumulate the gradient of supernet\par
    $\nabla\mathcal{W}_{\mathcal{A}_{t}} = get\_gradient (\mathcal{N}, \mathcal{W}, \mathcal{A}_{t}, train\_data)$,
    $\nabla{W} += \nabla\mathcal{W}_{\mathcal{A}_{t}}$
\ENDFOR
    \STATE Update the supernet weights $\mathcal{W} \leftarrow \nabla{W}$  
    \STATE Update current $Top\_T$ for next iteration $Top\_T$ = $Update\_Top\_T(Top\_T$, $\{\mathcal{A}_t\}_{t=1}^T$, $ \{ACC_{\mathcal{A}_t}\}_{t=1}^T)$
\UNTIL{The end of search stage}
\end{algorithmic}
\end{algorithm}

\subsection{Supernet Transferring}
The lack of transferring ability is a common problem for previous NAS methods. For a new dataset, most of the previous methods need to train a new supernet before searching stage. This is quite time-consuming, which may need several GPU days.

The success of transfer learning and pre-trained models has shown the inner-correlation of different downstream tasks and different datasets. Our method is inspired by fine-tuning, which is one of the most straight-forward and effective implementations of transfer learning.

With a little modification, our supernet shifting stage can be used to transfer a pre-trained supernet to a new dataset. Specifically, we keep the feature-extraction part of the supernet and only set a new fully-connected layer for prediction. The supernet is fine-tuned for tens of iterations when an architecture is sampled in the evolutionary searching algorithm. The loss is updated immediately instead of accumulating for faster transfer. Since the feature extraction part has a strong internal correlation on different datasets, only prediction layer needs to train from scratch. So the supernet is quickly shifted to a new dataset after several architectures are sampled. The searching algorithm can work properly.

We also try a different mode, freezing the feature extraction part and only fine-tuning the final prediction layer. In experiment, fine-tuning the whole supernet works better, especially on CIFAR-100 dataset. This is probably because ImageNet-1K dataset isn't large-scale enough. Thus, the general knowledge learned can't overcome the gap of data distribution. Moreover, since we adopt a lightweight search space, fine-tuning the whole supernet isn't time-consuming. If a more complicated supernet is pre-trained on a larger-scale dataset, the second mode may be better.

Note that the size of images may vary in different datasets, we try two different solutions. The first solution is to simply resize the image into a fixed size like $224\times224$, while the second solution is to set an up-sampling or down-sampling layer in the network. This layer is also initialized and trained during searching together with the prediction layer. By experiment, the former method works better.

In the experiment, we find the transferring supernet converges quickly after sampling several architecture. And after 4 iterations of evolutionary searching, the transferring supernet achieves nearly the same accuracy as a new supernet which is trained from scratch for 80,000 iterations on the new dataset. This strongly confirms that the supernet is theoretically transferable.

Note that our transferring method is time-saving with no loss of performance. Since no new supernet is trained, we make the searching process 10 times faster and even get a dominating architecture comparing with training a new supernet on ImageNet-100 dataset. The detailed experiments are shown in Sec.~\ref{sec:4.4}

\subsection{Approach Summary and Remarks}

\begin{table}[tb!]
  \centering
  \caption{Overview of different methods on improving relative performance estimation in weight-sharing strategy. Some works~\cite{fewshot} reduce the shared-weight. Some works~\cite{SUMNAS,NSAS} directly reduce gradient conflict. Some other works~\cite{greedy, attentive} including ours adopt a nonuniform sampling strategy to focus on superior architectures. Comparing with other methods, ours consumes little extra time, storage and introduce less early bias.}
  \vspace{-5pt}
  \large
  \resizebox{0.48\textwidth}{!}
{
  \begin{tabularx}{0.96\textwidth}{|>
  {\centering\arraybackslash}m{3.1cm}<{\centering}|>{\centering\arraybackslash}m{5.2cm}<{\centering}|>{\centering\arraybackslash}m{1.9cm}<{\centering}|>{\centering\arraybackslash}m{4.8cm}<{\centering}|}
  
    \hline
     \textbf{Method} & \textbf{Strategy} & \textbf{Extra storage} & \textbf{Extra time cost}\\
    \hline
    Few-shot NAS\cite{fewshot} & \textbf{Reduce shared weight: }split supernet into sub-supernets & sub-supernet & None\\
    \hline
    SUMNAS\cite{SUMNAS} & \textbf{Reduce gradient conflict: }Reducing gradient direction conflict in training & None &
    Compute reptile gradient in training \\
    \hline
    AttentiveNAS\cite{attentive} & \textbf{Nonuniform sample: }Focus on Pareto-best and worst in training & None & Pre-train evaluator + evaluation in training\\ 
    \hline
    GreedyNAS\cite{greedy} & \textbf{Nonuniform sample: }Multi-path sampling with rejection in training & None & Extra evaluation in training \\
    \hline
    
    Ours & \textbf{Nonuniform sample: }Shift supernet to focus on superior architectures in searching & None & Loss accumulate in searching(in total 2 GPU hours)\\
    
    \hline 
  \end{tabularx}
}
  \label{tab:compariosn}
\end{table}

We propose a refined searching method for NAS. By introducing supernet shifting, the supernet gradually focus on superior architectures and both global and local order-preserving ability can be ensured. Moreover, a pre-trained supernet can be transferred into other tasks effectively.

Table~\ref{tab:compariosn} shows the comparison of different method trying to overcome multi-model forgetting and improve performance estimation in weight-sharing. Ours consumes little extra time and storage. Encouraging supernet to focus on top architectures in searching instead of training further introduces less early bias, and supports transferability.

Our method also inherits the flexibility and simplicity of two-stage NAS. Multiple constraints can be added in the evolutionary searching stage to restrict the maximum flops, parameters and latency.

\section{Experiment}
\label{sec:4}
\subsection{Experiment Setting}

\textbf{Dataset.}  To show the general effect of our method, three different datasets are used. The biggest dataset is \emph{ImageNet-1K}~\cite{image}, which contains 1000 different classes, over one million images for training and 50,000 images for validation. \emph{ImageNet-1K} is the most important dataset in our experiment. 
Since getting retrain performance of a large number of architectures on \emph{ImageNet-1K} is time-consuming, we also use \emph{ImageNet-100} dataset. \emph{ImageNet-100} is a dataset sampled from \emph{ImageNet-1K}. It contains 100 classes and each class contains 1300 images for training and 50 images for validation. \emph{ImageNet-100} dataset is mainly used to evaluate the order-preserving ability and the performance of supernet transferring.
Besides, we also use \emph{Cifar-100}~\cite{cifar} in supplementary experiments.

\textbf{Search Space.}  Our search space also follows SPOS. It is based on ShuffleNet-V2~\cite{shuffle}, which is a powerful lightweight network. There are total 20 searching blocks and 4 choices for each block. The total search space is $4^{20}$. We use $FLOPS\leq330M$ as complexity constraint in evolutionary searching as well.

\textbf{Training.}  For the training of supernet and the retraining of the searched architectures, we use the same setting (including hyper-parameters, data-augmentation strategy, learning-rate decay, etc.) as SPOS. The batchsize is 1024, the supernet is trained for 150,000 iterations and the searched architecture is trained for 300,000 iterations on ImageNet-1K. For ImageNet-100, the batchsize is 256, the supernet is trained for 80,000 iterations and the searched architecture is trained for 120,000 iterations. To further ensure fairness, we use the same supernet weight for different searching algorithms if the search space is the same. Training uses 4 \textit{NVIDIA GeForce RTX 3090} GPUs and searching uses 1 \textit{NVIDIA GeForce RTX 3090} GPU.

\subsection{Searching Result}
\label{sec:4.2}

For comparison, we implement multiple searching algorithms and retrain the searched architecture. It involves:
 
1) Randomly select five architectures and choose the best architecture according to retrain performance

2) Oneshot(Train a supernet by uniform sampling and use random search to select the best architecture from 100 candidates according to supernet evaluation)

3) SPOS (train a supernet by uniform sampling and apply evolutionary searching algorithm)

4) FairNAS (train a supernet with strict-fairness sampling and apply evolutionary searching algorithm)
\begin{figure}[t!]
  \centering
   \includegraphics[width=\linewidth]{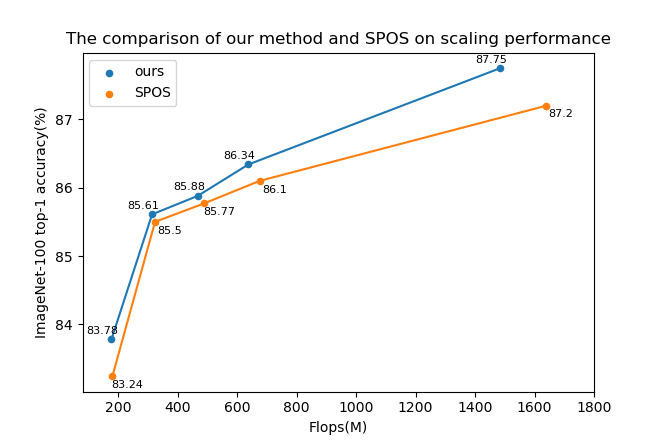}
\vspace{-15pt}
   \caption{We choose 5 depth multipliers: 0.5, 1.0, 1.5, 2.0 and 4.0. For each we train a new supernet to which we apply our method and the SPOS method. Then, we retrain the searched architecture separately and compare the results on ImageNet-100.}
   \label{fig:scale}
\end{figure}

Table~\ref{tab:result} shows the retrain result on different datasets of different NAS methods. We get a dominating architecture which has the lowest flops and the highest accuracy.

\begin{table}[tb!]
  \centering
  \caption{The retrain result on different datasets. For fairness, we implement different strategies ourselves in the same search space based on ShuffleNet-V2.}
  \vspace{-5pt}
  \resizebox{0.47\textwidth}{!}{
  \begin{tabular}{lccccc}
    \hline
    & \multicolumn{3}{c}{\textbf{ImageNet-1K}} & \multicolumn{2}{c}{\textbf{ImageNet-100}} \\
    \textbf{Method} & \textbf{Flops} & \textbf{Top-1 acc} & \textbf{Top-5 acc} & \textbf{Flops} & \textbf{Top-1 acc}\\
    \hline
    Random select & 324M & 73.29 & 91.01 & 312M & 85.42 \\
    Oneshot~\cite{oneshot}  & 326M & 73.52 & 91.44 & 304M & 85.41\\
    SPOS (block)~\cite{SPOS} & 323M & 74.01 & 92.25 & 304M & 85.50\\
    FairNAS~\cite{FAIR} & 326M & 74.03 & 92.31 & 300M & 85.44\\
    Ours & \textbf{318M} & \textbf{74.28} & \textbf{92.92} & \textbf{299M} & \textbf{85.61}\\
    \hline
  \end{tabular}}
  \label{tab:result}
\end{table}

To further verify the scaling performance, we use different depth multipliers to scale up the search space. By comparing with SPOS in Fig.~\ref{fig:scale}, our method outperforms SPOS under every depth multiplier.

Moreover, since most previous methods~\cite{attentive, FAIR, greedy} all make improvements in supernet training stage, our method can improve the searching quality by adding supernet shifting in evolutionary searching stage in a plug-and-play manner. The result is shown in Tab.~\ref{tab:new}
\begin{table}[t]
  \centering
  \caption{Retrain result on ImageNet-1K in different search spaces w/ or w/o supernet shifting for different baseline methods(implemented by ourselves).
  }
  \vspace{-5pt}
  \setlength{\tabcolsep}{1pt}
  \resizebox{0.48\textwidth}{!}{
  \begin{tabular}{l|c|c|c|c}
    \hline
    \multirow{2}{*}{\textbf{Method}} & \multicolumn{2}{c|}{\textbf{w/o shifting}}&\multicolumn{2}{c}{\textbf{w/ shifting}}  \\
    \cline{2-5}
    &\textbf{~Flops} (M)~ & \textbf{~Top-1 acc~} & \textbf{Flops} (M)& \textbf{Top-1 acc}\\
    \hline
    \multicolumn{5}{c}{ShuffleNet-V2 \textbf{(main search space)}}\\
    \hline
    SPOS & 323 & 74.01 & 318 ($\downarrow 1.5\%$) & 74.28 ($\uparrow 0.35\%$) \\ 
    FairNAS & 326 & 74.03 & 321 ($\downarrow 1.5\%$) & 74.36 ($\uparrow 0.45\%$)\\
    GreedyNAS & 329 & 74.17 & 325 ($\downarrow 1.2\%$) & 74.17 ($\rightarrow$)\\
    AttentiveNAS~ & 319 & 74.22 & 324 ($\uparrow 1.6\%$) & 74.38 ($\uparrow 0.22\%$)\\
    \hline
    \multicolumn{5}{c}{MobileNet-V2 \textbf{(supplementary search space)}}\\
    \hline
    SPOS & 333 & 73.42 & 329 ($\downarrow 1.2\%$) & 74.01 ($\uparrow 0.80\%$) \\
    FairNAS & 329 & 73.39 & 331 ($\uparrow 0.6\%$) & 74.13 ($\uparrow 1.01\%$) \\
    GreedyNAS & 336 & 73.59 & ~332 ($\downarrow 1.2\%$)~ & 73.82 ($\uparrow 0.31\%$)\\
    AttentiveNAS & 335 & 74.12 & 332($\downarrow 0.9\%$) & 74.23($\uparrow 0.15\%$) \\
    \hline
  \end{tabular}}
  \label{tab:new}
  \vspace{-10pt}
\end{table}

\subsection{Order-preserving Ability}
\label{sec:4.3}

As discussed before, for one-shot NAS problem using supernet as relative performance predictor, the most important feature of supernet is the order-preserving ability between supernet performance and retrain performance. Since it's time-consuming to get the retrain accuracy of a large number of architectures, we use ImageNet-100 in this part.

We analyze the global order-preserving ability and the local order-preserving ability separately. Specifically, We choose 10 good architectures and 20 random architectures as poor architectures. Each architecture is retrained from scratch for 50,000 iterations. We have verified that all 10 good architectures are better than the 20 poor architectures after retraining. Global order-preserving ability means whether the model is able to distinguish good architectures and poor architectures, so we counted how many of the top-10 architectures evaluated by supernet in the overall 30 architectures are the ten good architectures. And local order-preserving ability refers to whether the supernet can predict the relative performance of similar good architectures. So we introduce Kendall's tau coefficient, an index indicating the positive correlation of two ranks, as our metric. We calculate Kendall's tau coefficient of the 10 good architectures to evaluate the prediction of local relative performance. 

\begin{figure}[t!]
  \centering
   \includegraphics[width=\linewidth]{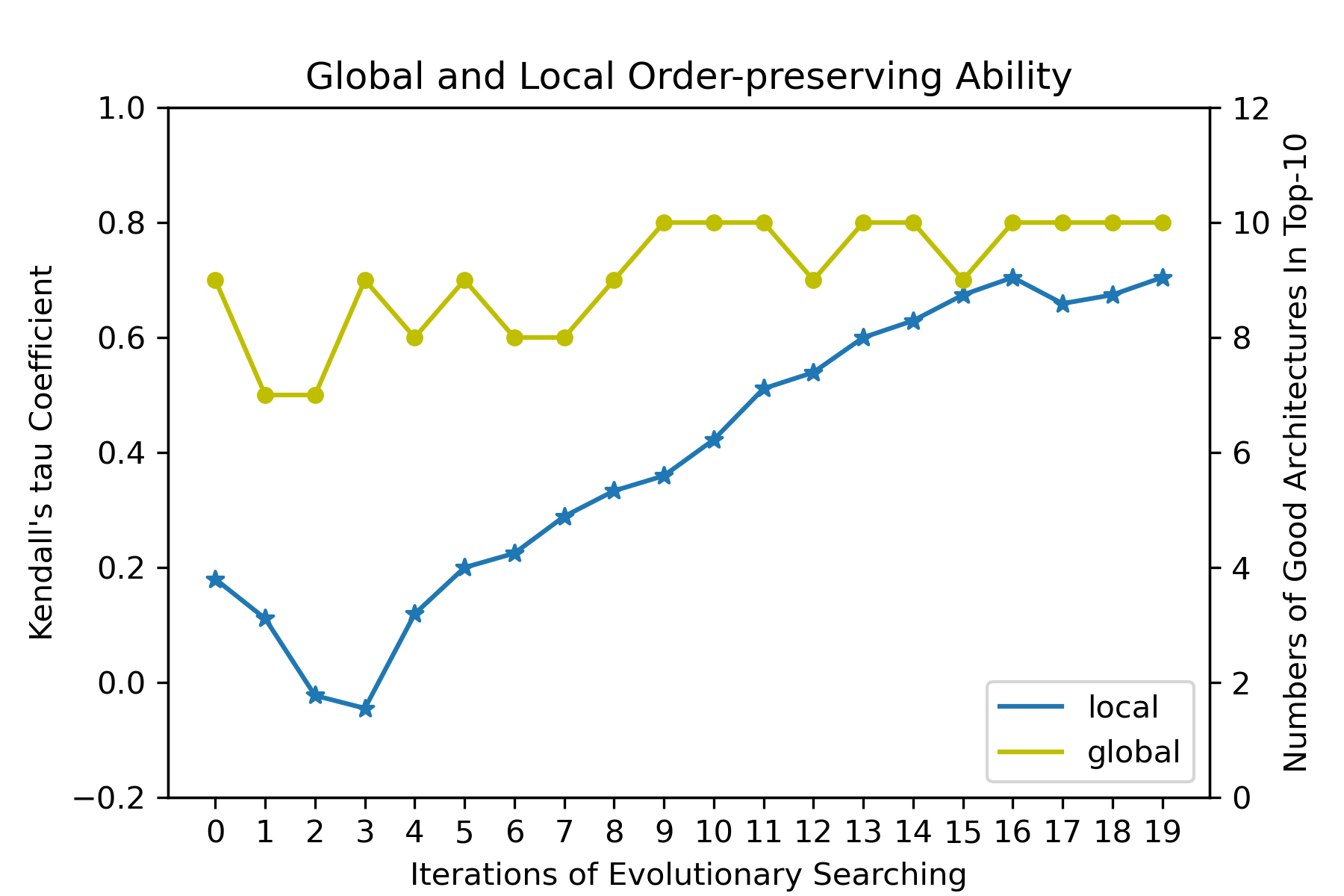}
 \vspace{-15pt}
   \caption{Experiments on order-preserving ability. The number of good architectures predicted correctly as the top-10 architectures indicates the global ranking. The Kendall's tau coefficient of the 10 good architectures indicates the local consistency.}
   \label{fig:order}
\end{figure}
Fig.~\ref{fig:order} shows the result of our experiment. We can see that for the supernet trained by uniform sampling strategy, the global ranking is quite ideal but the local consistency still needs to be improved. At the beginning of the searching stage, both the global ranking and the local consistency decreases. This is probably because the searching experience is not enough for evolutionary searching algorithm and thus the sampling doesn't satisfy Eq.~\ref{eq:71}. After several iterations of evolutionary algorithm, we can see a contiguous improving of local consistency and global ranking. This verifies that by using a reliable biased sample strategy like ours, the local order-preserving ability can be improved. After 15 iterations, the local Kendall's tau rate achieves a high score and remains nearly unchanged, which indicates the convergence, and ensures the fairness of final comparison.


\subsection{Supernet Transfer}
\label{sec:4.4}
To evaluate the transferability of our method, we first pre-train a supernet on ImageNet-1K and use our method to transfer the pre-trained supernet to downstream datasets including ImageNet-100 and Cifar100 during supernet shifting to search for the optimal architecture. We set three different method for comparison. 

1. \textbf{ImageNet-1K $\rightarrow$ downstream:} Choose the architecture searched on ImageNet-1K directly.

2. \textbf{downstream only - SPOS: } Train a new supernet and search by evolutionary algorithm on downstream dataset.

3. \textbf{downstream only - Ours:} Train a new supernet and search by evolutionary algorithm with supernet shifting on downstream dataset.

For these three different methods, the input size is all resized to $224\times224$ to ensure fairness.

Table~\ref{tab:resulttransfer} demonstrates the results. Our transferring method can speed up the search process for about 10 times by reusing the encoder of supernet. Meanwhile, the search quality does not decrease. Instead, we even find dominating architecture in ImageNet-100. This shows the supernet as pre-trained on large-scale datasets like ImageNet-1K contains general knowledge and can improve the performance on downstream datasets.

We find our transferring method usually prefers architectures with lower flops and fewer parameters. This is probably because simpler architectures usually converge faster and thus they can occupy an advantageous position at the start of evolutionary searching.

\begin{table}[tb!]
  \centering
  \caption{The retrain result on different transferring methods on ImageNet-100 and Cifar-100. Time cost consists of supernet training and searching and is measured in GPU hours.}
  \vspace{-5pt}
  \resizebox{0.48\textwidth}{!}{
  \begin{tabular}{lcccccc}
    \hline
    & \multicolumn{3}{c}{\textbf{ImageNet-100}} & \multicolumn{3}{c}{\textbf{CIFAR-100}}\\
    \textbf{Method} & \textbf{Flops} & \textbf{Top-1 acc} & \textbf{Time}& \textbf{Flops} & \textbf{Top-1 acc} & \textbf{Time}  \\
    \hline
    same architecture & 307M & 85.50 & / & 233M & 74.73 & / \\
    supernet & 304M & 85.50 & 48 & 228M & 75.40 & 20\\
    supernet + shifting & \textbf{299M} & 85.61 & 50 & 230M & \textbf{75.62} & 21\\
    Ours & 299M & \textbf{85.83} & 6 & \textbf{226M} & 75.38 & 3\\
    \hline
  \end{tabular}}
  \label{tab:resulttransfer}
\end{table}
\subsection{Time Cost Analysis}
As our method involves the supernet shifting together with evolutionary search, the additional time cost is rather small. Moreover, since supernet is fine-tuned in shifting, the requirements of the quality of pre-trained supernet is lower. We only need a rough global prediction on different architectures, and the local resolution can be improved during supernet shifting. Thus, the supernet training iterations can be reduced without significantly hurting performance.

In Table~\ref{tab:time}, we shorten supernet the training stage from 150K iterations to 100K iterations. Under the same flops, the top-1 accuracy searched by our method only decreases by 0.1, while the accuracy searched by SPOS drops by 0.4.

Therefore, under some resource-limited situations where the supernet cannot be trained for sufficient iterations and thus can't provide accurate estimation, our method can help to maintain performance.
\begin{table}[tb!]
  \centering
  \caption{Time cost on ImageNet-1K in GPU hours and the top-1 accuracy is estimated under the same Flops. The total search time is divided into supernet training time and evolutionary search time.}
   \vspace{-5pt}
   \resizebox{0.4\textwidth}{!}{
  \begin{tabular}{lccc}
    \hline
    \textbf{Method} & \textbf{Training} & \textbf{Search} & \textbf{Top-1 acc} \\
    \hline
    SPOS~\cite{SPOS} & 150 & 17 & 73.91 \\
    Ours & 150 & 20 & 74.11\\
    SPOS(short training)~\cite{SPOS} & 100  &17 & 73.52\\
    Ours(short training) & 100 & 20 & 73.96 \\
    \hline
  \end{tabular}}
  \label{tab:time}
\end{table}

\section{Conclusion and Outlook}
In this work, We have proposed supernet shifting, a strong and flexible method to improve order-preserving ability and transferability for Neural Architecture Search. With little additional time cost, the supernet can focus on superior architectures and thus ensure both local and global order-preserving ability. The search quality can be improved. Extensive experiments demonstrate the effectiveness.

Besides, a pre-trained supernet can be transferred to a new dataset in searching. This gives space for developing new NAS pipelines for future study. We believe a supernet pre-trained on a large-scale dataset will be open source and can be transferred to other tasks effectively. Optimal architectures can be searched by only applying searching algorithms with no need of tricky and time-consuming supernet training. This will significantly reduce the time cost, hardware restriction and searching difficulty for NAS. NAS and AutoML can be widely applied to various situations.

\textbf{Limitations.} 
For efficiency reasons, our supernet shifting is highly dependent on the sampling by EA. Although statistically correct, there are no explicit expressions on the sampling distribution. Mutation makes the sampling even more uncontrollable. Besides, whether it's fair and effective for different architectures to compare in an ever-changing supernet still needs to be verified.

{
    \small
    \bibliographystyle{ieeenat_fullname}
    \bibliography{main}
}

\input{sec/X_suppl}

\end{document}

%% file: preamble.tex
\usepackage[dvipsnames]{xcolor}


%% file: sec/X_suppl.tex


%



\title{Boosting Order-Preserving and Transferability for Neural Architecture Search: a Joint Architecture Refined Search and Fine-tuning Approach}



\clearpage
\setcounter{page}{1}
\maketitlesupplementary

\section{Search Space}

\begin{figure}[htb]
  \centering
   \includegraphics[width=0.98\linewidth]{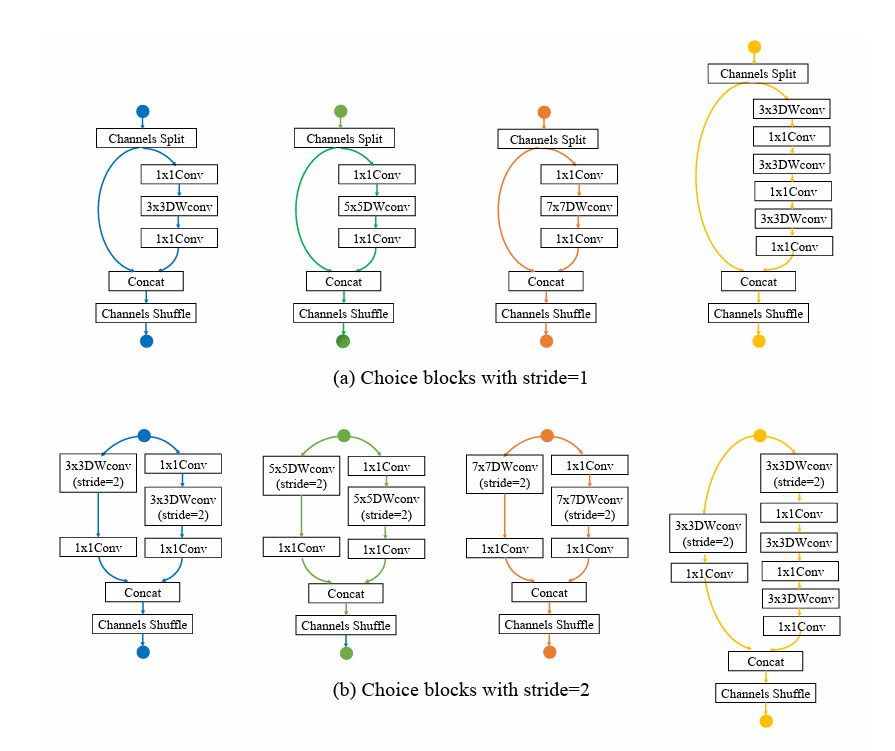}
   \vspace{-5pt}
   \caption{The choice block of our search space based on ShuffleNet-V2. From left to right: \textit{Choice\_3, Choice\_5, Choice\_7, Choice\_x}. The figure is the same as SPOS as we use the same search space.}
   \label{fig:choice}
\end{figure}
\begin{table}[htb]
  \centering
  \caption{Design of our supernet. CB: choice block. GAP: global average pooling. Stride: stride of the first layer in each block.}
  \vspace{-5pt}
    \resizebox{0.4\textwidth}{!}
{
  \begin{tabular}{c c c c c}
    \hline
    \textbf{Input size} & \textbf{Blocks} & \textbf{Channels} & \textbf{Repeat} & \textbf{Stride} \\
    \hline
    $224^{2}\times3$ & $3\times3$ conv &16 & 1 & 2\\
    $112^{2}\times16$ & CB & 64 & 4 & 2 \\
    $56^{2}\times64$ & CB & 160 & 4 & 2 \\
    $28^{2}\times160$ & CB & 320 & 8 & 2 \\
    $14^{2}\times320$ & CB & 640 & 4 & 2 \\
    $7^{2}\times640$ & $1\times1$conv & 1024 & 1 & 1 \\
    $7^{2}\times1024$ & GAP & 1024 & 1 & - \\
    1024 & fc & 1000 & 1 & - \\
    \hline
  \end{tabular}
  }
  \label{tab:supernet.}
\end{table}

Fig.~\ref{fig:choice} shows our choice block and search space, which is based on \textit{ShuffleNet-V2}, a strong lightweight convolutional neural network. Table~\ref{tab:supernet.} shows our supernet. The supernet helps embody our search space and provide a relative performance estimator for different architectures.

\section{Searching Result}
\begin{figure}[tb!]
  \centering
   \includegraphics[width=0.98\linewidth]{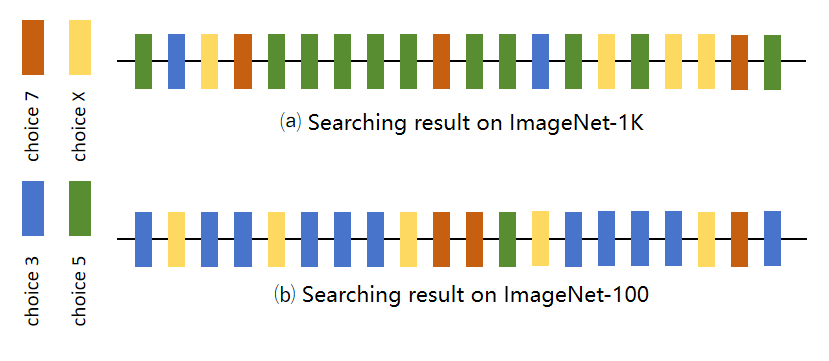}
   \vspace{-5pt}
   \caption{The searched architecture on ImageNet-1K and ImageNet-100}
   \label{fig:result}
\end{figure}
Fig.~\ref{fig:result} shows the searched architectures on ImageNet-1K and ImageNet-100 respectively. We can see that for ImageNet-100, it gets a rather simpler architecture in the same search space comparing with ImageNet-1K. This is in line with our intuition.

\section{Further discussion on Order-Preserving}
\label{sec:b}

We find that global order-preserving ability is relatively general among different tasks while local order-preserving ability is task-specific. In other words, some architectures are in-born superior and are more likely to perform better in different tasks. Our experimental results show that the top 10 architectures searched on ImageNet-1K outperform another 10 random architectures by 0.5\% on Cifar-100 and 0.9\% on ImageNet-100 on average. It implies global order can be roughly preserved across datasets. However, the ranking of the top 10 architectures by the performance of Cifar-100 and ImageNet-100 is quite different from the ranking by the performance of ImageNet-1K, showing that the local relative ranking is task-specific since it can perform differently in different tasks.

This is an interesting finding and it can provide a rough estimation and a better 'start point' for searching.

\section{Hyperparameter setting}
For the training of supernet and the retraining of the searched architectures, we use the same setting (including hyper-parameters, data-augmentation strategy, learning-rate decay, etc.) as SPOS. The batchsize is 1024, the supernet is trained for 150,000 iterations and the searched architecture is trained for 300,000 iterations on ImageNet-1K. For ImageNet-100, the batchsize is 256, the supernet is trained for 80,000 iterations and the searched architecture is trained for 120,000 iterations. 

The learning rate in supernet shifting is 1e-4. In total 640 samples are used to compute loss.

\section{Experiments on Edge Devices}
\begin{figure}[tb!]
  \centering
   \includegraphics[width=0.68\linewidth]{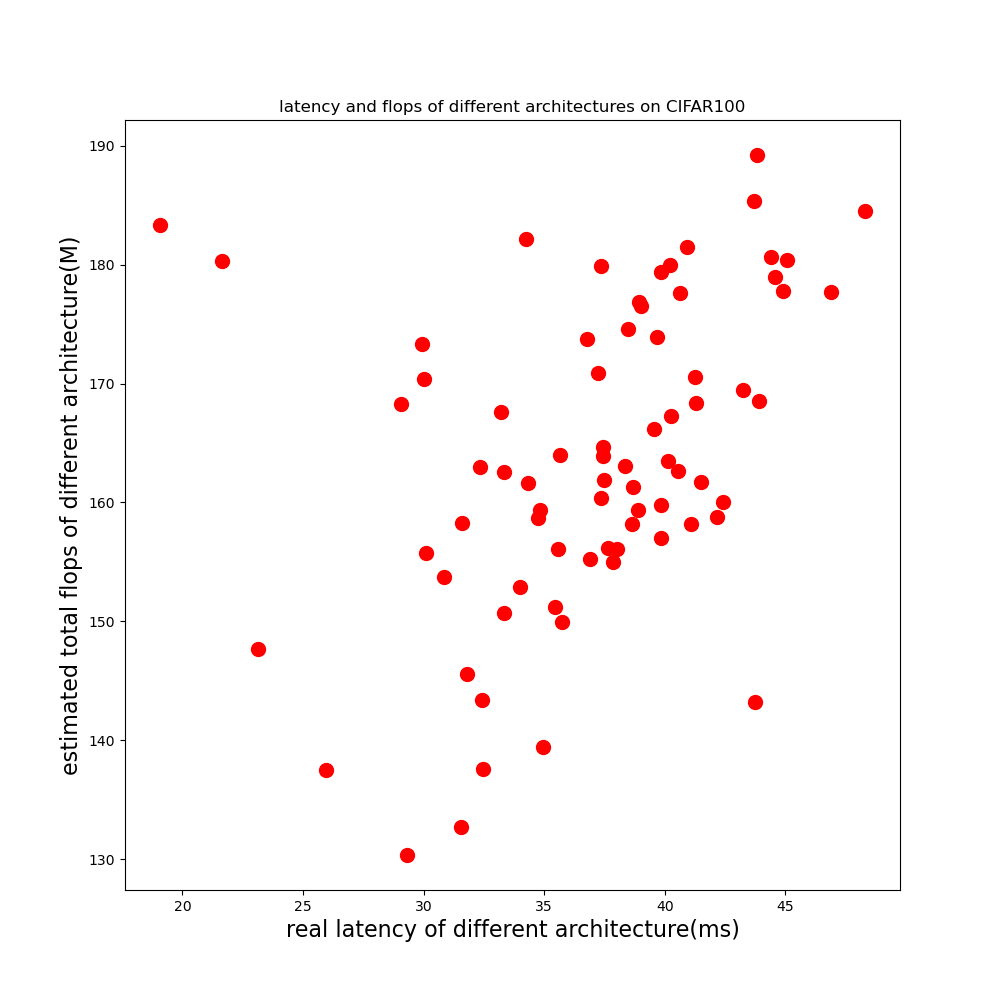}
   \vspace{-5pt}
   \caption{The correlation between flops and real-time latency on Cifar dataset.}
   \label{fig:latency}
\end{figure}
\begin{table}[tb!]
  \centering
  \caption{The retrain result of ImageNet-100 using different constraint in searching.}
  \begin{tabular}{cccc}
    \hline
    \textbf{Constraint} & \textbf{Top-1 acc} & \textbf{Flops} & \textbf{Latency} \\
    \hline
    Flops & 85.61 & 299M & 62.14ms \\
    Latency & 85.59 & 305M & 58.31ms\\
    \hline
  \end{tabular}
  \label{tab:latency-flops}
\end{table}
Unlike cloud servers, edge devices usually have constraints on the memory and computational resources, making it intractable to load complex models. Neural architecture search is one of the most popular and effective techniques that can design efficient neural architectures for edge devices with limited resources.

Our method can be easily applied to such an on-device NAS task.
Constraints on the resources, such as FLOPs and latency, can be seen as multiple objects in the evolutionary searching stage. Unlike most current NAS methods that use FLOPs to approximate the efficiency of architecture, our work also  supports utilizing the latency from real-time measurement on edge devices as one of the search objects.

Specifically, we apply our method on \textit{ROC-RK3588S-PC}, an 8-Core 8K AI Mainboard, and adopt the NSGA-II multiple object optimization method to search the Pareto superior to both accuracy and latency. The result is shown in Table~\ref{tab:latency-flops}. 

From the result, we can see that those architectures which have lower Flops do not necessarily have lower latency. Therefore, we further analyze the correlation between the FLOPs and latency on a large number of architectures. Results are shown in Fig.~\ref{fig:latency}. The Kendall's tau is only 0.3, which shows a non-negligible gap between Flops and real-time latency. Therefore, supporting real-time measurements is vital for applying NAS to on-device AI.

